\documentclass[journal,12pt,onecolumn,draftclsnofoot]{IEEEtran}
\usepackage[latin9]{luainputenc}
\usepackage{float}
\usepackage{amsmath}
\usepackage{amsthm}
\usepackage{graphicx}
\usepackage[unicode=true]
 {hyperref}

\makeatletter

\floatstyle{ruled}
\newfloat{algorithm}{tbp}{loa}
\providecommand{\algorithmname}{Algorithm}
\floatname{algorithm}{\protect\algorithmname}

\theoremstyle{plain}
\newtheorem{thm}{\protect\theoremname}
\theoremstyle{plain}
\newtheorem{lem}[thm]{\protect\lemmaname}
\theoremstyle{plain}
\newtheorem{prop}[thm]{\protect\propositionname}

\usepackage[caption=false,font=footnotesize]{subfig}
\usepackage{algorithm,algpseudocode}
\usepackage{cite}
\usepackage{amssymb}

\usepackage{graphicx}
\usepackage{epsfig}
\usepackage{epstopdf}
\usepackage{hyperref}

\makeatother

\providecommand{\lemmaname}{Lemma}
\providecommand{\propositionname}{Proposition}
\providecommand{\theoremname}{Theorem}

\begin{document}

\title{Sparse Reduced Rank Regression With Nonconvex Regularization}

\author{Ziping~Zhao,~\IEEEmembership{Student Member,~IEEE} and Daniel~P.~Palomar,~\IEEEmembership{Fellow,~IEEE}\thanks{This work was supported by the Hong Kong RGC 16208917 research grant.
The work of Z. Zhao was supported by the Hong Kong PhD Fellowship
Scheme (HKPFS).}\thanks{Z. Zhao and D. P. Palomar are with the Department of Electronic and
Computer Engineering, The Hong Kong University of Science and Technology
(HKUST), Clear Water Bay, Kowloon, Hong Kong (e-mail: \protect\href{mailto:ziping.zhao@connect.ust.hk}{ziping.zhao@connect.ust.hk};
\protect\href{mailto:palomar@ust.hk}{palomar@ust.hk}).}}
\maketitle
\begin{abstract}
In this paper, the estimation problem for sparse reduced rank regression
(SRRR) model is considered. The SRRR model is widely used for dimension
reduction and variable selection with applications in signal processing,
econometrics, etc. The problem is formulated to minimize the least
squares loss with a sparsity-inducing penalty considering an orthogonality
constraint. Convex sparsity-inducing functions have been used for
SRRR in literature. In this work, a nonconvex function is proposed
for better sparsity inducing. An efficient algorithm is developed
based on the alternating minimization (or projection) method to solve
the nonconvex optimization problem. Numerical simulations show that
the proposed algorithm is much more efficient compared to the benchmark
methods and the nonconvex function can result in a better estimation
accuracy.
\end{abstract}


\section{Introduction}

Reduced Rank Regression (RRR) \cite{Anderson1951,Anderson1984} is
a multivariate linear regression model, where the coefficient matrix
has a low-rank property. The name of ``reduced rank regression''
was first brought up by Izenman \cite{Izenman1975}. Denote the response
(or dependent) variables by $\mathbf{y}_{t}\in\mathbb{R}^{P}$ and
predictor (or independent) variables by $\mathbf{x}_{t}\in\mathbb{R}^{Q}$,
a general RRR model is given as follows:
\begin{equation}
\mathbf{y}_{t}=\boldsymbol{\mu}+\mathbf{A}\mathbf{B}^{T}\mathbf{x}_{t}+\boldsymbol{\varepsilon}_{t},\label{eq:RRR}
\end{equation}
where the regression parameters are $\boldsymbol{\mu}\in\mathbb{R}^{P}$,
$\mathbf{A}\in\mathbb{R}^{P\times r}$ and $\mathbf{B}\in\mathbb{R}^{Q\times r}$
and $\boldsymbol{\varepsilon}_{t}$ is the model innovation. Matrix
$\mathbf{A}$ is often called sensitivity (or exposure) matrix and
$\mathbf{B}$ is called factor matrix with the linear combinations
$\mathbf{B}^{T}\mathbf{x}_{t}$ called latent factors. The ``low-rank
structure'' formed by $\mathbf{A}\mathbf{B}^{T}$ essentially reduces
the parameter dimension and improves explanatory ability of the model.
The RRR model is widely used in situations when the response variables
are believed to depend on a few linear combinations of the predictor
variables, or when such linear combinations are of special interest. 

The RRR model has been used in many signal processing problems, e.g.,
array signal processing \cite{VibergStoicaOttersten1997}, state space
modeling \cite{StoicaJansson2000}, filter design \cite{MantonHua2001},
channel estimation and equalization for wireless communication \cite{LindskogTidestav1999,HuaNikpourStoica2001,NicoliSpagnolini2005},
etc. It is also widely applied in econometrics and financial economics.
Problems in econometrics were also the motivation for the pioneering
work on the RRR estimation problem \cite{Anderson1951}. In financial
economics, it can be used when modeling a group of economic indices
by the lagged values of a set of economic variables. It is also widely
used to model the relationship between financial asset returns and
some related explanatory variables. Several asset pricing theories
have been proposed for testing the efficiency of portfolios \cite{Zhou1995}
and empirical verification using asset returns data on industry portfolios
has been made through tests for reduced rank regression \cite{BekkerDobbelsteinWansbeek1996}.
The RRR model is also closely related the vector error correction
model \cite{ZhaoPalomar2017a} in time series modeling and the latent
factors can be used for statistical arbitrage \cite{ZhaoPalomar2018}
in finance. More applications on the RRR model can be found in, e.g.,
\cite{VeluReinsel2013}.

Like the low-rank structure for factor extraction, row-wise group
sparsity on matrix $\mathbf{B}$ can also be considered to further
realize predicting variable selection, which leads to the sparse RRR
(SRRR) model \cite{ChenHuang2012}. Since $\mathbf{B}^{T}\mathbf{x}_{t}$
can be interpreted as the linear factors linking the response variables
and the predictors, the SRRR can generate factors only with a subset
of all the predictors. Variable selection is very important target
in data analytics since it can help with model interpretability and
improve estimation and forecasting accuracy. 

In \cite{ChenHuang2012}, the authors first considered the SRRR estimation
problem, where the group sparsity was induced via the group lasso
penalty \cite{YuanLin2006}. An algorithm based on the alternating
minimization (AltMin) method \cite{Bertsekas1999} was proposed. However,
the proposed algorithm has a double loop where subgradient or variational
method is used for the inner problem solving. Such an algorithm can
be very slow in practice due to the double-loop nature where lots
of iterations may be necessary to get an accurate enough solution
at each iteration. Apart from that, besides the convex function for
sparsity inducing, it is generally acknowledged that a nonconvex sparsity-inducing
function can attain a better performance \cite{FanLi2001} which is
proposed to use for sparsity estimation in this paper.

In this paper, the objective of the SRRR estimation problem is given
as the ordinary least squares loss with a sparsity-inducing penalty.
An orthogonality constraint is added for model identification purpose
\cite{ChenHuang2012}. To solve this problem, an efficient AltMin-based
single-loop algorithm is proposed. In order to pursue low-cost updating
steps, a majorization-minimization method \cite{SunBabuPalomar2016}
and a nonconvexity redistribution method \cite{YaoKwok2018} are further
adopted making the variable updates become two closed-form projections.
Numerical simulations show that the proposed algorithm is more efficient
compared to the benchmarks and the nonconvex function can attain a
better estimation accuracy. 

\section{Sparse Reduced Rank Regression}

The SRRR estimation problem is formulated as follows:
\begin{equation}
\begin{array}{ll}
\underset{\mathbf{A},\mathbf{B}}{\mathsf{minimize}} & F\left(\mathbf{A},\mathbf{B}\right)\triangleq L\left(\mathbf{A},\mathbf{B}\right)+R\left(\mathbf{B}\right)\\
\mathsf{subject\:to} & \mathbf{A}^{T}\mathbf{A}=\mathbf{I},
\end{array}\label{eq:Problem Formulation}
\end{equation}
where $L\left(\mathbf{A},\mathbf{B}\right)$ is sample loss function
and  $R\left(\mathbf{B}\right)$ is the row-wise group sparsity regularizer.
The constraint $\mathbf{A}^{T}\mathbf{A}=\mathbf{I}$ is added for
identification purpose to deal with the unitary invariance of the
parameters \cite{ChenHuang2012}. We further assume a sample path
$\left\{ \mathbf{y}_{t},\mathbf{x}_{t}\right\} _{t=1}^{N}$ $\left(N\geq\max\left(P,Q\right)\right)$
is available from \eqref{eq:RRR}. 

The least squares loss $L\left(\mathbf{A},\mathbf{B}\right)$ for
the RRR model is obtained by minimizing a sample $\ell_{2}$-norm
loss as follows\footnote{In this paper, the intercept term has been omitted without loss of
generality as in \cite{ChenHuang2012}, since it can always be removed
by assuming that the response and predictor variables have zero mean.}:
\begin{equation}
\begin{array}{c}
L\left(\mathbf{A},\mathbf{B}\right)=\frac{1}{2}\sum_{t=1}^{N}\left\Vert \mathbf{y}_{t}-\mathbf{A}\mathbf{B}^{T}\mathbf{x}_{t}\right\Vert _{2}^{2}\\
\quad\quad=\frac{1}{2}\left\Vert \mathbf{Y}-\mathbf{A}\mathbf{B}^{T}\mathbf{X}\right\Vert _{F}^{2},
\end{array}\label{eq:LS loss}
\end{equation}
where $\mathbf{Y}=\left[\mathbf{y}_{1},\ldots,\mathbf{y}_{N}\right]$
and $\mathbf{X}=\left[\mathbf{x}_{1},\ldots,\mathbf{x}_{N}\right]$. 

\begin{figure}
\centering{}\includegraphics[scale=0.6]{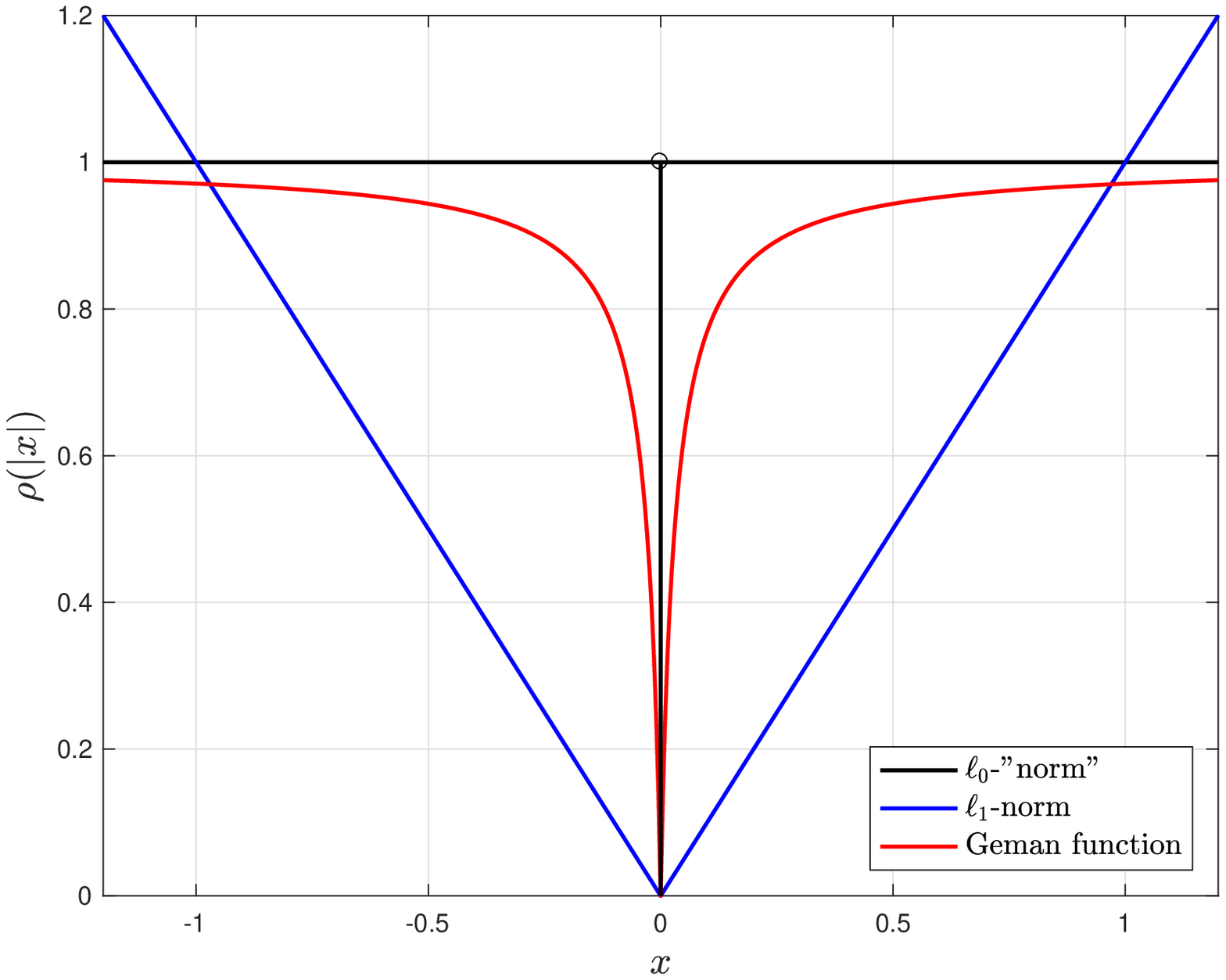}\caption{\label{fig:nonsmooth-sparsity-inducing-functions}The $\ell_{0}$-''norm''
$\left\Vert x\right\Vert _{0}\triangleq\mathrm{sgn}\left(x\right)$,
$\ell_{1}$-norm $\rho_{\ell_{1}}\left(\left|x\right|\right)\triangleq\left|x\right|$,
and nonconvex nonsmooth sparsity-inducing function $\rho\left(\left|x\right|\right)$.}
\end{figure}

Sparse optimization \cite{BachJenattonMairalObozinskiothers2012}
has become the focus of much research interest as a way to to realize
the variable selection (e.g., the group lasso method). For a vector
$\mathbf{x}\in\mathbb{R}^{K}$, the sparsity level is usually measured
by the $\ell_{0}$-norm, i.e., $\left\Vert \mathbf{x}\right\Vert _{0}=\sum_{i=1}^{K}\mathrm{sgn}\left(\left|x_{i}\right|\right)$.
Practically, the $\ell_{1}$-norm is used as the tightest convex relaxation
to approximate it as in \cite{ChenHuang2012}. Although it is easy
for optimization and has been shown to favor sparser solutions, the
$\ell_{1}$-norm can lead to biased estimation with solutions not
as accurate and sparse as desired and produce inferior prediction
performance \cite{FanLi2001}. Nonconvex regularizers sacrifice convexity
but can have a tighter approximation performance and are proposed
for sparsity inducing which outperform the convex $\ell_{1}$-norm.
In this paper, two nonsmooth sparsity-inducing functions denoted by
$\rho\left(\left|x\right|\right)$ are considered: the nonconvex Geman
function \cite{GemanReynolds1992} and the convex $\ell_{1}$-norm.
Then, the row-wise group sparsity regularizer $R\left(\mathbf{B}\right)$
induced by $\rho\left(\left|x\right|\right)$ is given as follows:
\begin{equation}
\begin{array}{c}
R\left(\mathbf{B}\right)=\sum_{i=1}^{Q}\xi_{i}\rho\left(\left\Vert \mathbf{b}_{i}\right\Vert _{2}\right),\end{array}\label{eq:group-wise sparsity}
\end{equation}
where $\mathbf{b}_{i}$ denotes the $i$th row of $\mathbf{B}$ and
$\rho\left(\left|x\right|\right)$ is from $\rho_{\mathrm{GM}}\left(\left|x\right|\right)=\frac{\left|x\right|}{\theta+\left|x\right|}$
($\theta>0$) and $\rho_{\ell_{1}}\left(\left|x\right|\right)=\left|x\right|$,
which are shown in Figure \ref{fig:nonsmooth-sparsity-inducing-functions}.

Based on $L\left(\mathbf{A},\mathbf{B}\right)$ and $R\left(\mathbf{B}\right)$,
the problem in \eqref{eq:Problem Formulation} is a nonconvex nonsmooth
optimization problem due to the nonconvex nonsmooth objective and
the nonconvex constraint set.

\section{Problem Solving Based on Alternating Minimization}

The objective function in problem \eqref{eq:Problem Formulation}
has two variable blocks $\left(\mathbf{A},\mathbf{B}\right)$. In
this section, an alternating minimization (a.k.a. two-block coordinate
descent) algorithm \cite{Bertsekas1999} will be proposed to solve
it. At the $\left(k+1\right)$th iteration, this algorithm updates
the variables according to the following two steps:
\begin{equation}
\begin{cases}
\begin{array}{l}
\mathbf{A}^{\left(k+1\right)}\leftarrow\arg\underset{\mathbf{A}:\mathbf{A}^{T}\mathbf{A}=\mathbf{I}}{\min}F\left(\mathbf{A};\mathbf{B}^{\left(k\right)}\right)\\
\mathbf{B}^{\left(k+1\right)}\leftarrow\arg\underset{\mathbf{B}}{\min}F\left(\mathbf{B};\mathbf{A}^{\left(k+1\right)}\right),
\end{array}\end{cases}
\end{equation}
where $\left(\mathbf{A}^{\left(k\right)},\mathbf{B}^{\left(k\right)}\right)$
are updates generated at the $k$th iteration.

First, let us start with the minimization step w.r.t. variable $\mathbf{A}$
when $\mathbf{B}$ is fixed at $\mathbf{B}^{\left(k\right)}$, the
problem becomes\footnote{For simplicity, $F\left(\mathbf{A};\mathbf{B}^{\left(k\right)}\right)$
is written as $F\left(\mathbf{A}\right)$ and likewise the fixed variables
$\mathbf{A}^{\left(k\right)}$ and/or $\mathbf{B}^{\left(k\right)}$
in other functions will also be reduced in the following.}
\begin{equation}
\begin{array}{ll}
\underset{\mathbf{A}}{\mathsf{minimize}} & F\left(\mathbf{A}\right)\simeq\frac{1}{2}\left\Vert \mathbf{Y}-\mathbf{A}\mathbf{B}^{\left(k\right)T}\mathbf{X}\right\Vert _{F}^{2}\\
\mathsf{subject\:to} & \mathbf{A}^{T}\mathbf{A}=\mathbf{I},
\end{array}\label{eq:LSE A}
\end{equation}
where the ``$\simeq$'' means ``equivalence'' up to additive constants.
This nonconvex problem is the classical orthogonal Procrustes problem
(projection) \cite{GowerDijksterhuis2004}, which has a closed-form
solution given in the following lemma.
\begin{lem}
\label{lem:orthogonal-Procrustes-problem}\cite{GowerDijksterhuis2004}
The orthogonal Procrustes problem in \eqref{eq:LSE A} can be equivalently
reformulated into the following form:
\[
\begin{array}{ll}
\underset{\mathbf{A}}{\mathsf{minimize}} & \left\Vert \mathbf{A}-\mathbf{P}_{A}^{\left(k\right)}\right\Vert _{F}^{2}\\
\mathsf{subject\:to} & \mathbf{A}^{T}\mathbf{A}=\mathbf{I},
\end{array}
\]
where $\mathbf{P}_{A}^{\left(k\right)}\triangleq\mathbf{Y}\mathbf{X}^{T}\mathbf{B}^{\left(k\right)}$\textup{.}
Let the thin singular value decomposition (SVD) be $\mathbf{P}_{A}=\mathbf{U}\mathbf{S}\mathbf{V}^{T}$,
where $\mathbf{U}\in\mathbb{R}^{Q\times r}$ and $\mathbf{S},\mathbf{V}\in\mathbb{R}^{r\times r}$,
then the optimal update $\mathbf{A}^{\left(k+1\right)}$ is given
by
\begin{equation}
\ensuremath{\mathbf{A}}^{\left(k+1\right)}=\mathbf{U}\mathbf{V}^{T}.\label{eq:BM A}
\end{equation}
\end{lem}
Then, when fixing $\mathbf{A}$ with $\ensuremath{\mathbf{A}}^{\left(k+1\right)}$,
the problem for $\mathbf{B}$ is
\begin{equation}
\begin{array}{ll}
\underset{\mathbf{B}}{\mathsf{minimize}} & F\left(\mathbf{B}\right)=\frac{1}{2}\left\Vert \mathbf{Y}-\mathbf{A}^{\left(k+1\right)}\mathbf{B}^{T}\mathbf{X}\right\Vert _{F}^{2}\\
 & \quad\quad\quad\quad+\sum_{i=1}^{Q}\xi_{i}\rho\left(\left\Vert \mathbf{b}_{i}\right\Vert _{2}\right),
\end{array}\label{eq:LSE B}
\end{equation}
which is a penalized multivariate regression problem. It has no analytical
solution but standard nonconvex optimization algorithms or solvers
can be applied to solve it. However, using such methods will lead
to an iterative process, which could be undesirable in terms of efficiency.
In addition, since the nonconvexity of this problem, if no guarantee
for the solution quality can be claimed, the overall alternating algorithm
in general is not guaranteed to converge to a meaningful point.

In this paper, the $\mathbf{B}$-subproblem is solved via a simple
update rule while guaranteeing convergence of the overall algorithm.
We propose to update $\mathbf{B}$ by solving a majorized surrogate
problem for problem \eqref{eq:LSE B} \cite{SunBabuPalomar2016,HongRazaviyaynLuoPang2016}
written as
\begin{equation}
\mathbf{B}^{\left(k+1\right)}\leftarrow\arg\underset{\mathbf{B}}{\min}\overline{F}\left(\mathbf{B};\mathbf{A}^{\left(k+1\right)},\mathbf{B}^{\left(k\right)}\right),
\end{equation}
where $\overline{F}\left(\mathbf{B};\mathbf{A}^{\left(k+1\right)},\mathbf{B}^{\left(k\right)}\right)$
or simply $\overline{F}\left(\mathbf{B}\right)$ is the majorizing
function of $F\left(\mathbf{B}\right)$ at ($\mathbf{A}^{\left(k+1\right)},\mathbf{B}^{\left(k\right)}$).
To get $\overline{F}\left(\mathbf{B}\right)$, we need the following
results.
\begin{lem}
\label{prop:quadratic majorization }\cite{SunBabuPalomar2016} Let
$\mathbf{A}\in\mathbb{S}^{K}$, then at any point $\mathbf{x}^{\left(k\right)}\in\mathbb{R}^{K}$,
\textbf{$\mathbf{x}^{T}\mathbf{A}\mathbf{x}$} is majorized as follows:
\[
\begin{aligned}\mathbf{x}^{T}\mathbf{A}\mathbf{x}\leq & \mathbf{x}^{\left(k\right)T}\mathbf{A}\mathbf{x}^{\left(k\right)}+2\mathbf{x}^{\left(k\right)T}\mathbf{A}\left(\mathbf{x}-\mathbf{x}^{\left(k\right)}\right)\\
 & +\psi\left(\mathbf{A}\right)\left\Vert \mathbf{x}-\mathbf{x}^{\left(k\right)}\right\Vert _{2}^{2},
\end{aligned}
\]
where $\psi\left(\mathbf{A}\right)\geq\lambda_{\max}\left(\mathbf{A}\right)$
is a pre-specified constant.
\end{lem}
Observing that the first part in $F\left(\mathbf{B};\mathbf{A}^{\left(k+1\right)}\right)$,
i.e., the least squares loss $L\left(\mathbf{B};\mathbf{A}^{\left(k+1\right)}\right)$,
is quadratic in $\mathbf{B}$, based on Proposition \ref{prop:quadratic majorization },
we can have the following result.
\begin{lem}
\label{lem:majorization-L(B)}The function $L\left(\mathbf{B};\mathbf{A}^{\left(k+1\right)}\right)$
can be majorized at $\left(\mathbf{A}^{\left(k+1\right)},\mathbf{B}^{\left(k\right)}\right)$
by
\[
\overline{L}\left(\mathbf{B}\right)\simeq\frac{1}{2}\psi(\mathbf{G}^{\left(k\right)})\left\Vert \mathbf{B}-\mathbf{P}_{B}^{\left(k\right)}\right\Vert _{F}^{2},
\]
where $\mathbf{G}^{\left(k\right)}\triangleq\mathbf{A}^{\left(k+1\right)T}\mathbf{A}^{\left(k+1\right)}\otimes\mathbf{X}\mathbf{X}^{T}$,
$\psi\left(\mathbf{G}^{\left(k\right)}\right)\geq\lambda_{\max}\left(\mathbf{G}^{\left(k\right)}\right)$,
and $\mathbf{P}_{B}^{\left(k\right)}\triangleq\psi^{-1}\left(\mathbf{G}^{\left(k\right)}\right)\mathbf{X}\mathbf{Y}^{T}\mathbf{A}^{\left(k+1\right)}-\psi^{-1}\left(\mathbf{G}^{\left(k\right)}\right)\mathbf{X}\mathbf{X}^{T}\mathbf{B}^{\left(k\right)}\mathbf{A}^{\left(k+1\right)T}\mathbf{A}^{\left(k+1\right)}+\mathbf{B}^{\left(k\right)}$.
\end{lem}
\begin{IEEEproof}
The proof is trivial and hence omitted.
\end{IEEEproof}
Likewise, the majorization method can also be applied to the regularizer
$R\left(\mathbf{B}\right)$. But we first need the following result.
\begin{prop}
\label{prop:nonconvexity redistribution}\cite{YaoKwok2018} The nonsmooth
sparsity-inducing function $\rho\left(\left|x\right|\right)$ can
be decomposed as
\[
\rho\left(\left|x\right|\right)=\kappa\left|x\right|+\rho\left(\left|x\right|\right)-\kappa\left|x\right|,
\]
where $\rho\left(\left|x\right|\right)-\kappa\left|x\right|$ is a
smooth and concave function when $\kappa\triangleq\rho^{\prime}\left(0^{+}\right)$.
Specifically, for $\rho_{\ell_{1}}\left(\left|x\right|\right)$, $\kappa=1$;
and for $\rho_{\mathrm{GM}}\left(\left|x\right|\right)$, $\kappa=1/\theta$.
\end{prop}
\begin{figure}
\centering{}\includegraphics[scale=0.6]{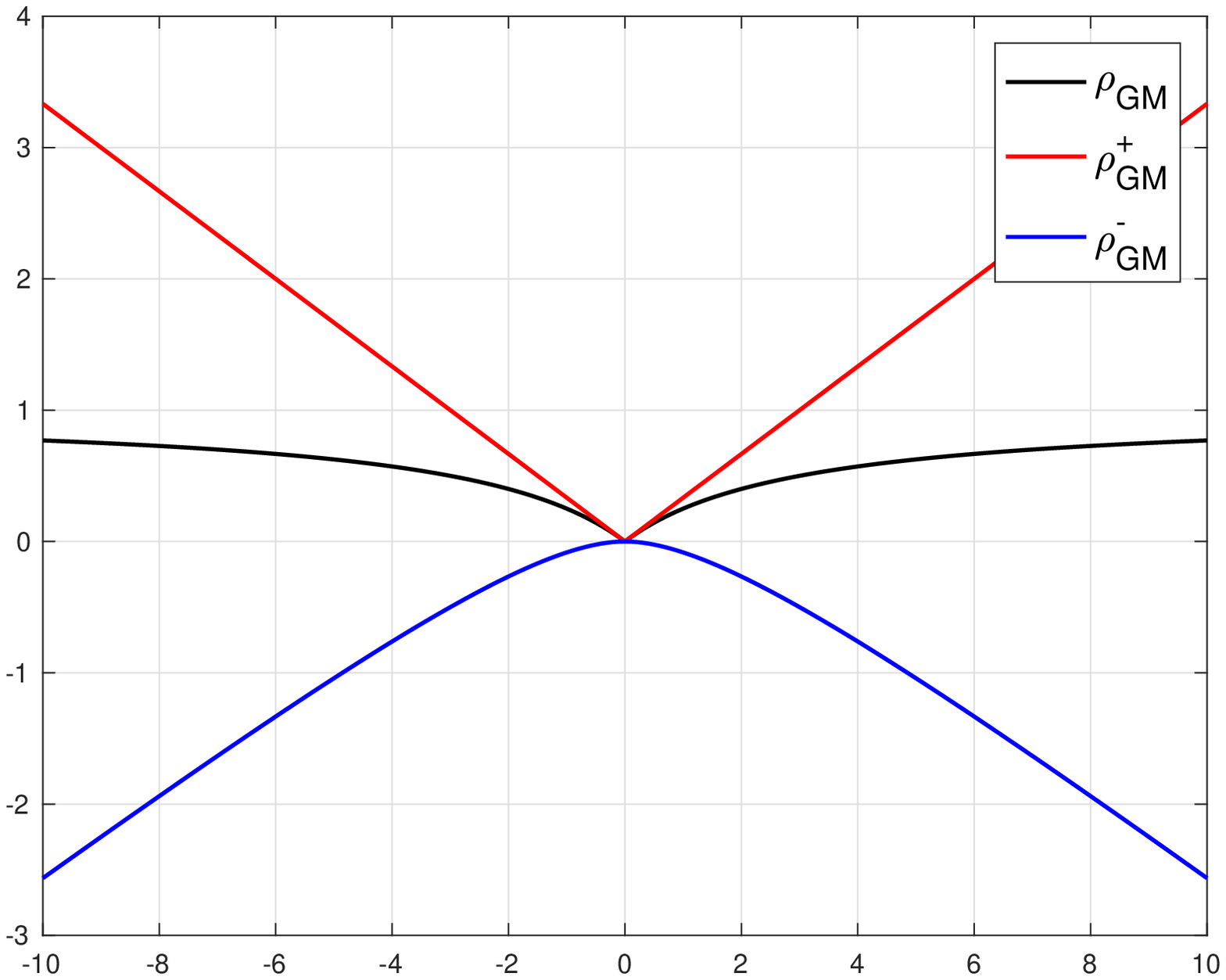}\caption{\label{fig:Nonconvexity-Redistribution}Nonconvexity Redistribution
Method for $\rho_{\mathrm{GM}}\left(\left|x\right|\right)$}
\end{figure}

An illustrating example for Proposition \ref{prop:nonconvexity redistribution}
is given in Figure \ref{fig:Nonconvexity-Redistribution}. And based
on Proposition \ref{prop:nonconvexity redistribution}, we can accordingly
decompose the row-wise group sparsity regularizer $R\left(\mathbf{B}\right)$
as
\begin{equation}
R\left(\mathbf{B}\right)=R^{+}\left(\mathbf{B}\right)+R^{-}\left(\mathbf{B}\right),
\end{equation}
where $R^{+}\left(\mathbf{B}\right)=\kappa\sum_{i=1}^{Q}\xi_{i}\left\Vert \mathbf{b}_{i}\right\Vert _{2}$
which exactly takes the form of classical group lasso and $R^{-}\left(\mathbf{B}\right)=R\left(\mathbf{B}\right)-R^{+}\left(\mathbf{B}\right)$.
For $R^{-}\left(\mathbf{B}\right)$, we can have the following majorization
result.
\begin{lem}
\label{lem:majorization-R(B)}The function $R^{-}\left(\mathbf{B}\right)$
can be majorized at $\mathbf{B}^{\left(k\right)}$ by
\[
\overline{R}^{-}\left(\mathbf{B}\right)\simeq\mathrm{Tr}\left(\mathbf{K}^{\left(k\right)T}\mathbf{B}\right),
\]
where $\mathbf{K}^{\left(k\right)}=R^{-\prime}\left(\mathbf{B}^{\left(k\right)}\right)$
with $R^{-\prime}\left(\mathbf{B}^{\left(k\right)}\right)$ to be
the gradient of $R^{-}\left(\mathbf{B}\right)$ at point $\mathbf{B}^{\left(k\right)}$
and specifically
\[
\mathbf{k}_{i}^{\left(k\right)}\triangleq\xi_{i}\left[\rho^{\prime}\left(\left\Vert \mathbf{b}_{i}^{\left(k\right)}\right\Vert _{2}\right)-\kappa\right]\frac{\mathbf{b}_{i}^{\left(k\right)}}{\left\Vert \mathbf{b}_{i}^{\left(k\right)}\right\Vert _{2}},
\]
where $\mathbf{k}_{i}^{\left(k\right)}$ denotes the $i$th column
of $\mathbf{K}^{\left(k\right)}$.
\end{lem}
\begin{IEEEproof}
The proof is trivial and hence omitted.
\end{IEEEproof}
Based on $\overline{L}\left(\mathbf{B}\right)$ in Lemma \ref{lem:majorization-L(B)}
and $\overline{R}^{-}\left(\mathbf{B}\right)$ in Lemma \ref{lem:majorization-R(B)},
we can finally have the majorization function for $F\left(\mathbf{B}\right)$
given as
\begin{equation}
\begin{array}{l}
\overline{F}\left(\mathbf{B}\right)=\overline{L}\left(\mathbf{B}\right)+R^{+}\left(\mathbf{B}\right)+\overline{R}^{-}\left(\mathbf{B}\right)\\
\quad\quad\quad\simeq\frac{1}{2}\psi\left(\mathbf{G}^{\left(k\right)}\right)\left\Vert \mathbf{B}-\mathbf{P}_{B,R}^{\left(k\right)}\right\Vert _{F}^{2}+R^{+}\left(\mathbf{B}\right),
\end{array}
\end{equation}
where $\mathbf{P}_{B,R}^{\left(k\right)}\triangleq\mathbf{P}_{B}^{\left(k\right)}-\psi^{-1}\left(\mathbf{G}^{\left(k\right)}\right)\mathbf{K}^{\left(k\right)}$.
The result by using Lemma \ref{lem:majorization-L(B)} and Lemma \ref{lem:majorization-R(B)}
is that we shift the nonconvexity associated with the nonconvex regularizer
to the loss function, and transform the nonconvex regularizer to the
familiar convex group lasso regularizer. It is easy to observe that
the algorithm derivation above can be easily applied to the classical
group lasso and at that case $\mathbf{K}^{\left(k\right)}=\mathbf{0}$. 

Finally, the majorizing problem for the $\mathbf{B}$-subproblem is
given in the following form:
\begin{equation}
\begin{array}{cl}
\underset{\mathbf{B}}{\mathsf{minimize}} & \frac{1}{2}\psi\left(\mathbf{G}^{\left(k\right)}\right)\left\Vert \mathbf{B}-\mathbf{P}_{B,R}^{\left(k\right)}\right\Vert _{F}^{2}+\frac{1}{\theta}\sum_{i=1}^{Q}\xi_{i}\left\Vert \mathbf{b}_{i}\right\Vert _{2}\end{array},\label{eq:subprobelm-B-MM-2}
\end{equation}
which becomes separable among the rows of matrix $\mathbf{B}$. The
resulting separable problems can be efficiently solved using the proximal
algorithms \cite{ParikhBoydothers2014} and have closed-form solutions
which are given in the following lemma.
\begin{lem}
\label{lem:proximal} \cite{ParikhBoydothers2014} The problem in
\eqref{eq:subprobelm-B-MM-2} has a closed-form proximal update which
is given by
\[
\begin{array}{l}
\mathbf{b}_{i}^{\left(k+1\right)}=\left[1-\frac{1}{\theta}\frac{\xi_{i}}{\psi\left(\mathbf{G}^{\left(k\right)}\right)}\frac{1}{\left\Vert \mathbf{p}_{i}^{\left(k\right)}\right\Vert _{2}}\right]^{+}\mathbf{p}_{i}^{\left(k\right)},\end{array}
\]
where $\left[x\right]^{+}\triangleq\max\left(x,0\right)$, and $\mathbf{p}_{i}^{\left(k\right)}$
is the $i$th row of $\mathbf{P}_{B,R}^{\left(k\right)}$.
\end{lem}

\subsection{AltMin-MM: Algorithm for SRRR Estimation}

Based on the alternating minimization algorithm together with the
majorization and nonconvex redistribution methods, to solve the original
SRRR estimation problem \eqref{eq:Problem Formulation}, we just need
to update the variables with closed-form solutions alternatingly until
convergence. 

The overall algorithm is summarized in the following.
\begin{center}
\begin{algorithm}[h]
\begin{algorithmic}[1]
\Require $\mathbf{X}$, $\mathbf{Y}$ and $\xi_{i}$ with $i=1,\ldots,r$.
\State Set   $k=0$, $\mathbf{A}^{(0)}$ and $\mathbf{B}^{(0)}$.

\Repeat

\State Compute $\mathbf{P}_{A}^{(k)}$

\State Update $\mathbf{A}^{\left(k+1\right)}$ in closed-form solution (Lemma \ref{lem:orthogonal-Procrustes-problem})

\State Compute $\mathbf{G}^{(k)}$, $\psi(\mathbf{G}^{(k)})$ and $\mathbf{P}_{B,R}^{(k)}$

\State Update $\mathbf{B}^{\left(k+1\right)}$ in closed-form solution (Lemma \ref{lem:proximal})
		
\State $k\gets k+1$

\Until convergence

\end{algorithmic}\caption{AltMin-MM: Algorithm for SRRR Estimation}
\end{algorithm}
\par\end{center}

\section{Numerical Simulations\label{sec:Numerical-Simulations}}

In order to test the performance of the problem model and proposed
algorithm. Numerical simulations are considered in this section. An
SRRR $\left(P=7,\:Q=5,\:r=3\right)$ with underlying group sparse
structure for $\mathbf{B}$ is specified firstly. Then a sample path
$\left\{ \mathbf{x}_{t},\mathbf{y}_{t},\boldsymbol{\varepsilon}_{t}\right\} _{t=1}^{N}$
is generated. 

\begin{figure}[h]
\begin{centering}
\includegraphics[width=0.6\columnwidth]{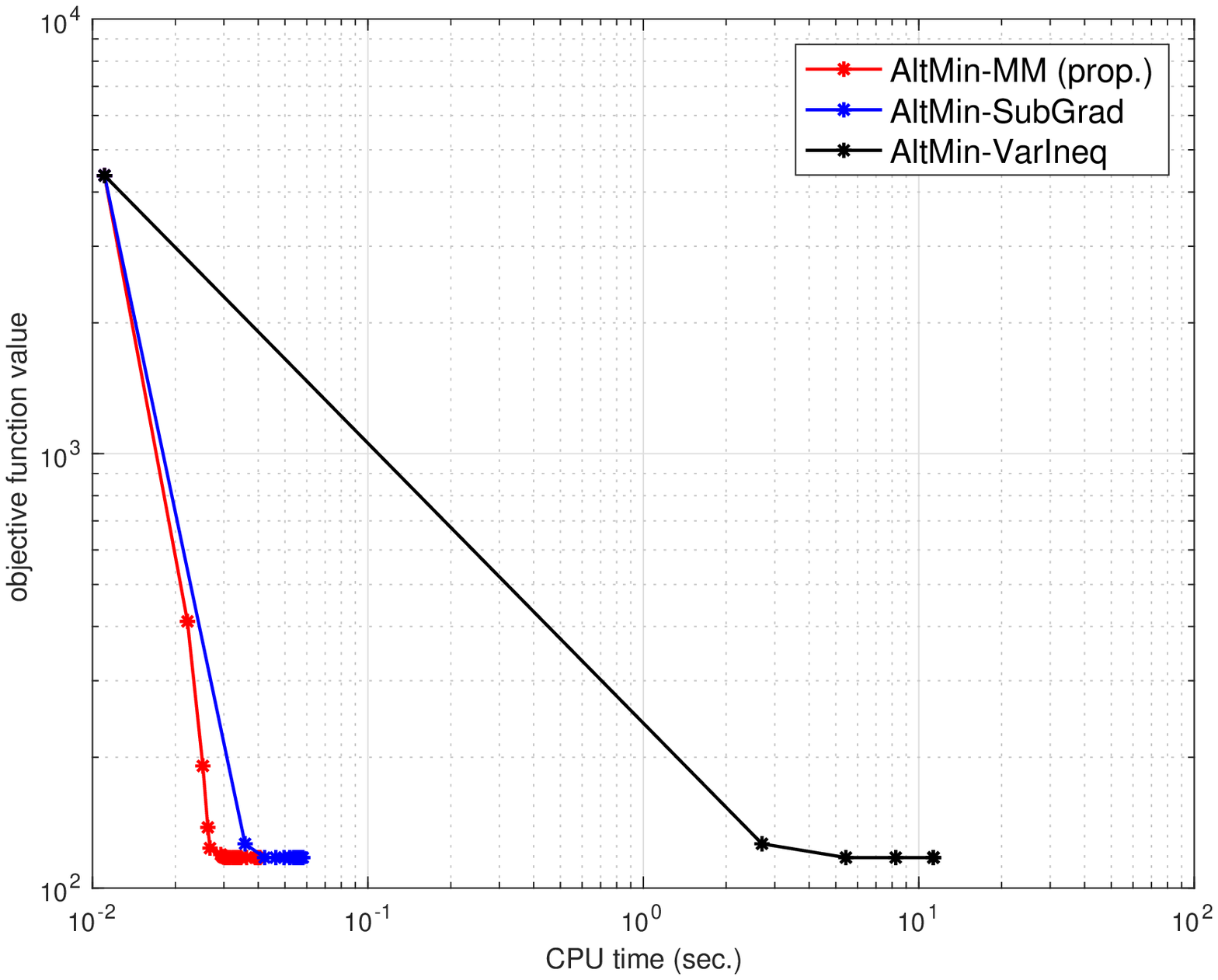}
\par\end{centering}
\centering{}\caption{\label{fig:Convergence-comparison}Convergence comparison for objective
function value ($N=100$).}
\end{figure}

We first examine the efficiency of our proposed AltMin-MM algorithm
when the sparsity regularizer is the group lasso penalty, i.e., $\rho\left(\left|x\right|\right)=\left|x\right|$
which is adopted in \cite{ChenHuang2012}. We compare our algorithm
with the AltMin-based algorithms with subproblem solved by subgradient
method (AltMin-SubGrad) and by variational inequality method (AltMin-VarIneq)
for the proposed problem in \eqref{eq:Problem Formulation}. The convergence
result of the objective function value is shown in Fig. \ref{fig:Convergence-comparison}.
It is easy to see that our proposed algorithm can have a faster convergence.
It should be mentioned that although the first descent step can attain
a better solution in the benchmark methods, since a lot of iterations
can be required to get a accuracy enough solution, they show a slower
convergence in general. 

\begin{figure}[h]
\begin{centering}
\includegraphics[width=0.6\columnwidth]{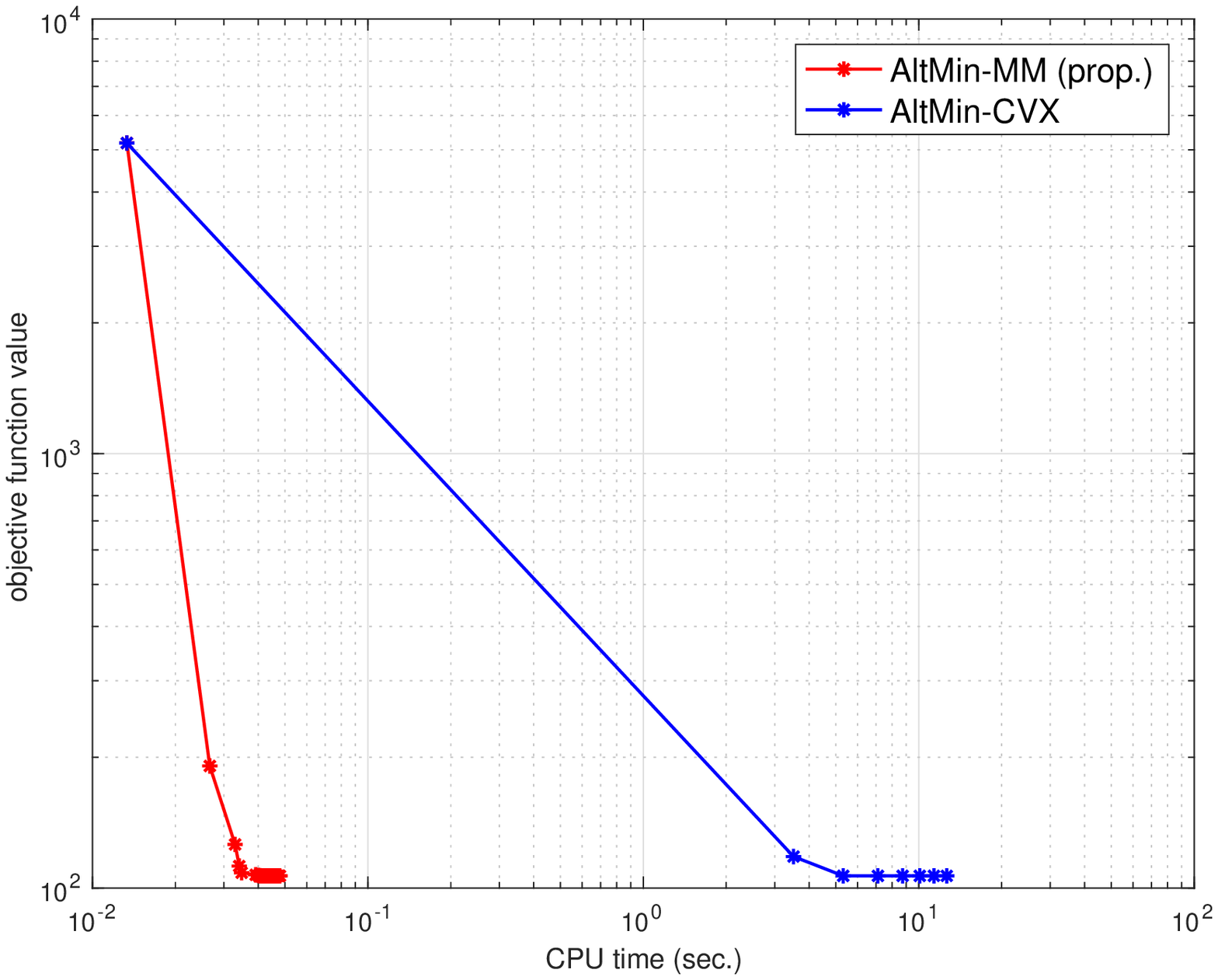}
\par\end{centering}
\centering{}\caption{\label{fig:Convergence-comparison-1}Convergence comparison for objective
function value ($N=100$).}
\end{figure}

We further test the case when the regularizer is based on nonconvex
Geman function, i.e., $\rho\left(\left|x\right|\right)=\frac{\left|x\right|}{\theta+\left|x\right|}$
($\theta=0.05$). Since there is no benchmark in the literature, our
proposed algorithm AltMin-MM is compared with a benchmark where the
convex $\mathbf{B}$-subproblem is derived to be a tight majorized
problem of the original problem by just majorizing the nonconvex term
$R^{-}\left(\mathbf{B}\right)$ and is solved using $\mathtt{CVX}$.
The objective function convergence result is shown in Fig. \ref{fig:Convergence-comparison}
and Fig. \ref{fig:Convergence-comparison-1}.

We also examine the estimation accuracy of the proposed formulation
and algorithm. It is evaluated by computing the angle between the
estimated factor matrix space $\hat{\mathbf{B}}^{\left(m\right)}$
and the true space $\mathbf{B}$ denoted by $\theta^{\left(m\right)}(\hat{\mathbf{B}}^{\left(m\right)},\mathbf{B})$
for the $m$th Monte-Carlo simulation, with $m=1,\ldots,M$ and $M=500$.
The angle $\theta^{\left(m\right)}(\hat{\mathbf{B}}^{\left(m\right)},\mathbf{B})$
is computed as follows \cite{Anderson1984}. First, compute the QR
decompositions $\hat{\mathbf{B}}^{\left(m\right)}=\mathbf{Q}_{m}\mathbf{R}_{m}$
and $\mathbf{B}=\mathbf{Q}\mathbf{R}$. Next, compute the SVD of $\mathbf{Q}_{m}^{T}\mathbf{Q}=\mathbf{U}_{Q}\mathbf{S}_{Q}\mathbf{V}_{Q}$
where the diagonal elements of $\mathbf{S}_{Q}$ is written as $s_{1}\geq\ldots\geq s_{r}$.
Then, the minimum angle is given by $\theta^{\left(m\right)}(\hat{\mathbf{B}}^{\left(m\right)},\mathbf{B})=\arccos\left(s_{1}\right)$.
The averaged angle for $M$ Monte-Carlo runs is given by
\[
\begin{array}{c}
\theta(\hat{\mathbf{B}},\mathbf{B})=\frac{1}{M}\sum_{m=1}^{M}\theta^{\left(m\right)}(\hat{\mathbf{B}}^{\left(m\right)},\mathbf{B}),\end{array}
\]
where it can take values from $0$ (identical subspaces) to $\frac{\pi}{2}$
(orthogonal subspaces). We compared three cases which are RRR estimation
(without sparsity), SRRR estimation with convex sparsity-inducing
function $\rho_{\ell_{1}}\left(\left|x\right|\right)$, and SRRR estimation
with nonconvex sparsity-inducing function $\rho_{\mathrm{GM}}\left(\left|x\right|\right)$.
It is easy to say that, the SRRR problem formulation can really exploit
the group sparsity structure in $\mathbf{B}$ and the nonconvex function
$\rho_{\mathrm{GM}}\left(\left|x\right|\right)$ shows a better performance
over the convex one.

\begin{figure}[h]
\centering{}\includegraphics[scale=0.6]{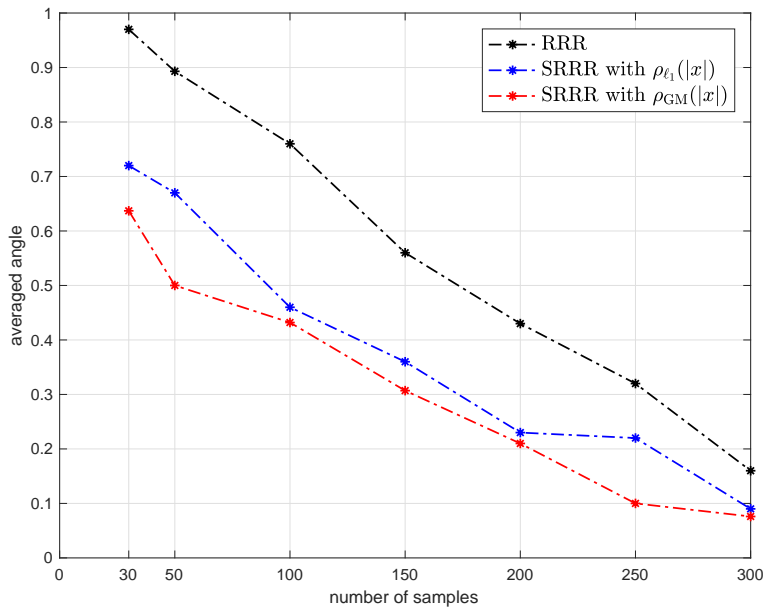}\caption{\label{fig:MR-Var}Estimation accuracy based on averaged angle.}
\end{figure}

\section{Conclusions \label{sec:Conclusions}}

The SRRR model estimation problem has been considered in this paper.
It has been formulated to minimize the least squares loss with a group
sparsity penalty and considering an orthogonality constraint. A nonconvex
nonsmooth sparsity function has been proposed. Efficient algorithm
based on the alternating minimization method, the majorization-minimization
method and the nonconvexity redistribution method has been developed
with variables updated in closed-form. Numerical simulations have
shown that the proposed algorithm is more efficient compared to the
benchmarks and the nonconvex regularizer can result in a better performance
than the convex one.

 \bibliographystyle{IEEEtran}
\bibliography{/Users/ziping/Dropbox/Research/1-Report/Reference/RefAll}

\end{document}